\documentclass[11pt,english,leqno]{article}
\usepackage[T1]{fontenc}
\usepackage[utf8]{inputenc}
\usepackage{tabto}
\usepackage{refstyle}
\usepackage{graphicx}
\usepackage{booktabs}
\graphicspath{ {./figures/} }
\usepackage{amsmath}
\usepackage{amssymb}
\usepackage{amsthm}
\usepackage{microtype}
\usepackage{tabularx}
\usepackage{makecell}
\usepackage{soul}
\usepackage{adjustbox}

\makeatletter

\AtBeginDocument{}
\AtBeginDocument{}
\AtBeginDocument{}

\RS@ifundefined{subsecref}
  {\newref{subsec}{name = \RSsectxt}}
  {}
\RS@ifundefined{thmref}
  {\def\RSthmtxt{theorem~}\newref{thm}{name = \RSthmtxt}}
  {}
\RS@ifundefined{lemref}
  {\def\RSlemtxt{lemma~}\newref{lem}{name = \RSlemtxt}}
  {}

\usepackage[bibtex]{LeL}
\usepackage{ifxetex}
\ifxetex
\usepackage{unicode-math}
\usepackage{fontspec}
\usepackage{polyglossia}
\fi

\usepackage{gb4e}
\noautomath

\renewcommand{\glt}{\small\trans}
\def\leltitlefs{\fontsize{17}{21}\selectfont}

\newcommand{\specialcell}[2][c]{%
  \begin{tabular}[#1]{@{}l@{}}#2\end{tabular}}

\makeatother

\usepackage{babel}

\begin{document}

\title{BAMBI: Developing BAby language Models for Italian}
\author{Alice Suozzi~~~~~Luca Capone\\~~~~~Gianluca E. Lebani~~~~~Alessandro Lenci} 
\maketitle

\begin{abstract}

This paper presents BAMBI (BAby language Models Boostrapped for Italian), a series of Baby Language Models (BabyLMs) trained on data that mimic the linguistic input received by a five-years-old Italian-speaking child. The BAMBI models are tested using BaBIEs \citep{capone2024babies}, a benchmark specifically designed to evaluate LMs, which takes into account the amount of training input the models received. The BAMBI models are compared against a large language model (LLM) and a multimodal language model (VLM), to study the contribution of extralinguitic information for language acqsuisition. The results of our evaluation align with the existing literature on English LMs, confirming that while reduced training data support the development of relatively robust syntactic competence, they are insufficient for fostering semantic understanding. However, the gap between the training resources (data and computation) of the BAMBI models and the LLMs is not fully reflected in their performance: Despite LLMs' massive training, their performance is not much better than that of BAMBI models. This suggests that strategies beyond scaling training resources, such as data curation, inclusion of multimodal input, and other training strategies (such as curriculum learning), could play a crucial role in shaping models' performance.\footnotemark
\\

\footnotetext{For the specific purposes of Italian Academy, Alice Suozzi for Sections 4.1, 4.2 and 4.3, Luca Capone is responsible for Sections 3.2, 4.4, Gianluca E. Lebani for Sections 2 and 3.1, and Alessandro Lenci for Section 1 and 5.}

\end{abstract}


\section{Introduction}
This paper introduces \textbf{BAMBI}, a series of BabyLMs (Baby Language Models) trained on data designed to be both qualitatively and quantitatively cognitively plausible. Specifically, we focus on linguistic input equivalent in size to that of a five-year-old, sourced from appropriate materials, such as transcriptions of speech directed at children. These models represent the initial outcome of the CLEVER (Computational and Linguistic bEnchmarks for the study of VErb argument structuRe) project, a larger research initiative, which pursues two main objectives. On the scientific side, the project explores training strategies and investigates the learning processes of LMs compared to human language acquisition. Specifically, it seeks to determine whether a training corpus that more closely mirrors children language experience enhances model training. From a technical point of view, CLEVER aims at  developing more sustainable and efficient LMs by leveraging small-scale, accessible resources. Here, we present the initial results obtained with BAMBI, which were evaluated using a benchmark specifically designed to assess the learning capabilities of Italian BabyLMs, BaBIEs \citep{capone2024babies}. Our findings appear to align with existing evidence from the literature on English LMs, which suggests that limited pretraining is sufficient to develop strong syntactic competence, whereas more training data and computation are required for effectively addressing semantically related tasks \citep{warstadt2023findings, hu2024findings}. 

The paper is structured as follows: Section \ref{sect_relatedworks} provides a brief introduction to the topic of LMs' cognitive plausibility, with a particular focus on data in shaping models' language learning. 
In Section \ref{sect_bambi}, the models are presented, highlighting the training data and the key characteristics of the BabyLMs \ref{subsect_modelsandtraining}. The remaining parts of the paper are devoted to describing the evaluation of such models, in comparison to two Large LMs, Minerva and SmolVLM-Instruct. The benchmark and the metrics used for the evaluation are overviewed in Sections \ref{subsect_babies} and \ref{subsect_metrics}, while the results are detailed and discussed in Sections \ref{subsect_results} and \ref{subsect_discussion}. Finally, some conclusions are drawn in Section \ref{sect_conclusions}. 

\section{A quick dive into LMs' cognitive plausibilty}\label{sect_relatedworks}
The cognitive implausibility of Large LMs is a recurring topic in the Natural Language Processing (NLP) literature. Numerous authors highlighted semantic \citep{bisk2020experience, bender2020climbing, merrill2021provable}, syntactic \citep{pater2019generative, dupre2021can, zhou2023well}, and cognitive \citep{bishop2021artificial, borji2023categorical, katzir2023large, mahowald2024dissociating} shortcomings in LMs. These criticisms are countered by an equally substantial body of responses. Regarding grounding and a priori constraints, several authors argue that linguistic systems can articulate sophisticated semantic content without relying on grounding \citep{gastaldi2021can, capone2021theory, abdou2021can, piantadosi2022meaning, sogaard2022understanding, sogaard2023grounding, patel2022mapping}. Others highlight the ability of models to acquire the grammar of natural-historical languages as a by-product of pre-training \citep{goldberg2019assessing, linzen2021syntactic, piantadosi2023modern}, framing this as one of many emerging capacities exhibited by LMs \citep{wei2022emergent}. Some even suggest that, if certain plausibility criteria are met, LMs could serve as useful models for studying language and its acquisition \citep{warstadt2022artificial, connell2024can, lenci2023understanding, cai2024large}. In the debate, none of the contributions appear conclusive. It seems plausible that LMs are capable of autonomously organizing semantic and syntactic content leveraging only next token prediction. At the same time, the functioning of these systems does not rule out the possibility that incorporating features of human ontogenetic development into pretraining strategies could improve model learning. 

A key aspect in child language acquisition, and the focus of the present work, is the quantity and quality of linguistic input. The increasing size of datasets, while advancing model performance, raises concerns about environmental sustainability, availability of training resources, and limitations for the future development of LMs \citep{villalobos2022will}. Additionally, this growth highlights a mismatch between the linguistic input received by models and that received by humans, complicating efforts to draw general conclusions about language acquisition and cognitive development. From a purely quantitative perspective, datasets contain several orders of magnitude more words than the linguistic input to which a human being is exposed \citep{warstadt2022artificial}: \cite{Lan:etal:2024} estimated that the training data for ChatGPT correspond to $36,540$ person-years. From a qualitative perspective, the training data of LMs are primarily composed of web-derived written texts. In contrast, human language learning involves much smaller yet richer and more diverse multimodal inputs, embedded within social and contextual environments. Even focusing on the linguistic data only, the input humans, and especially children, are exposed to exhibits distinct characteristics over time. These include variations in average sentence length, richness of vocabulary, progressive complexity of topics, and the tenor and pace of interactions \citep{hart1996meaningful, greenwood2011assessing, montag2018quantity, tal2023speakers}. Building on these considerations, this work presents four LMs trained on a dataset designed to more closely resemble the linguistic input to which Italian-speaking children are typically exposed.

\section{BAMBI: our Baby(LM)} \label{sect_bambi}
\subsection{Dataset} \label{subsect_dataset}
The models presented here are trained on the first portion of a larger set of data planned for the CLEVER project. This subset is designed to simulate the linguistic input to which an Italian-speaking 5-years-old child is typically exposed. According to the literature, children encounter an average of 10 million words per year \citep{hart1996meaningful, greenwood2011assessing, hu2024findings}. Over the first five years, this amounts to approximately 50 million words. The training dataset contains roughly 25 million words, presented to the model twice, resulting in a cumulative exposure of approximately 50 million words. The dataset only includes transcriptions of oral texts, so as to more closely resemble the kind of input received by children up to 5 years of age. The data are carefully selected from specific sources. Approximately, 1 million words consist of transcriptions of Child-Directed Speech. These include words from the CHILDES corpus \citep{macwhinney1984childes, sanchez2019childes}, and transcripts from studies on child language acquisition \citep{longobardi2015children, whittle2015insegnamento, spinelli2023there}. The remaining words are sourced from transcripts of real-life interactions, as well as educational and entertainment multimedia content targeted at children in the relevant age group (e.g., cartoons, movies, educational TV shows, etc.).

\subsection{Models and training} \label{subsect_modelsandtraining}
Two types of models are selected for training: encoder-only (a RoBERTa based model \citep{liu2019roberta}) and decoder-only (a GPT-2 based model \citep{radford2019language}) architectures (see Table \ref{tab:arch-param}). These architectures were chosen because they are the standard ones for BabyLMs training.

The models are trained using the HuggingFace Trainer,\footnote{\url{https://huggingface.co/docs/transformers/v4.48.0/main_classes/trainer}} following the specifications outlined in Table \ref{tab:train-param}.
In the BabyLM community \citep{warstadt2023findings, hu2024findings} there is typically no established limit on the number of epochs on which a model can be trained. Consequently, a dataset with a limited number of elements can be presented to the model repeatedly an indefinite number of times. However, this approach does not provide a means to evaluate how much a model can learn based solely on the linguistic input a preschooler would receive. To address this, two models are trained for each architecture (see Table \ref{tab:train-param}): one for two epochs and another continuing until the performance stops improving (patience value set to 3), for a maximum of 40 epochs. This setup ensures that the model trained for two epochs more closely simulates the linguistic experience of a child, while the unrestricted model serves as a useful comparison. The training strategy is intentionally kept standard and consistent across models to isolate the effect of the dataset on a regular LM architecture.

\begin{table}[!h]
\centering
\resizebox{\textwidth}{!}{ 
\begin{tabular}{@{}lcccc@{}}
\toprule
\textbf{Hyperparameter}          & \textbf{decoder} & \textbf{encoder} & \textbf{Minerva-3b-base-v1.0} & \textbf{SmolVLM-Instruct} \\ \midrule
Vocab size                       & 30.000           & 30.000            & 32.768            & 49.155            \\
Max length                       & 1024              & 512               & 16.384            & 16.384         \\
Hidden size                      & 768             & 256               & 2.560             & 2.048        \\
Attention heads                            & 12               & 8                 & 32                & 32         \\
Layers                           & 12               & 6                 & 32                & 24          \\
Trainable params                 & 131,922,432      & 26,630,704        & 2.894.236.160     & 2.246.272.880   \\
\bottomrule
\end{tabular}
}
\caption{Models Hyperparameters}
\label{tab:arch-param}
\end{table}

\begin{table}[!h]
\centering
\resizebox{\textwidth}{!}{ 
\begin{tabular}{@{}lcccc@{}}
\toprule
\textbf{Argument}          & 
\specialcell[t]{\textbf{2e\_train} \\ \textbf{\_decoder}} & 
\specialcell[t]{\textbf{full\_train} \\ \textbf{\_decoder}} & 
\specialcell[t]{\textbf{2e\_train} \\ \textbf{\_encoder}} & 
\specialcell[t]{\textbf{full\_train} \\ \textbf{\_encoder}} \\ \midrule
Initial learning rate   & 5e-4  & 5e-4  & 5e-4  & 5e-4  \\
Batch size              & 32    & 32    & 32    & 32    \\
Maximum epochs          & \textbf{2} & \textbf{40} & \textbf{2} & \textbf{40}   \\
Training epochs         & \textbf{2} & \textbf{40} & \textbf{2} & \textbf{15}   \\
Early stopping patience & \textbf{//} & \textbf{3} & \textbf{//} & \textbf{3}   \\
Grad. accumulation steps & 8    & 8     & 8     & 8     \\
lr scheduler type       & cosine & cosine & cosine & cosine \\
Warmup steps            & 1000  & 1000  & 1000  & 1000  \\
Weight decay            & 0.01  & 0.01  & 0.01  & 0.01  \\
fp16                    & True  & True  & True  & True  \\       
Metric                  & Loss  & Loss  & Loss  & Loss  \\    
\bottomrule
\end{tabular}
}
\caption{Training arguments}
\label{tab:train-param}
\end{table}

Due to differences in size, the unrestricted models completed different numbers of epochs before reaching a plateau. By the end of training, the unrestricted models processed 1 billion and 375 million words, respectively. The larger model (decoder) completed all 40 training epochs, achieving a final loss of 2.01, while the smaller encoder ended training after 15 epochs, with a loss of 20.24. In comparison, the restricted models recorded a final loss of 3.29 for the decoder and 24.89 for the encoder.

For evaluation purposes, the four trained models are compared with two additional pre-trained models. Minerva-3b-base-v1.0\footnote{\url{https://huggingface.co/sapienzanlp/Minerva-3B-base-v1.0}} and SmolVLM-Instruct,\footnote{\url{https://huggingface.co/HuggingFaceTB/SmolVLM-Instruct}} hereafter referred to as Minerva and SmolVLM. Minerva is selected for its being a nature Italian LM, while SmolVLM is chosen for its multimodal training with visual data. Although multimodal models showed limited impact in both the first and the second BabyLM Challenges \citep{warstadt2023findings, hu2024findings}, it is worth investigating whether this trend persists on a benchmark specifically designed for children in the early stages of language acquisition. Specifically, it is interesting to determine whether a multimodal VLM can outperform BabyLMs and a size-comparable LLM in tasks originally designed for assessing the linguistic abilities of Italian-speaking children (cf. Section \ref{subsect_babies}), for whom language acquisition  heavily relies on multimodality.

\section{Evaluating the models}

\subsection{BaBIEs: a Benchmark for the Linguistic Evaluation of Italian Baby Language Models} \label{subsect_babies}

BaBIEs \citep{capone2024babies} is a benchmark specifically created to evaluate the linguistic skills of BabyLMs in Italian. This tool, whose structure is summarized in Table \ref{tab:babies_structure}, consists of 419 items grouped into five different tasks. All items are adapted from four standardized tests designed to assess the linguistic abilities of Italian-speaking children. 

\begin{table}[!h]
\centering
\resizebox{\textwidth}{!}{ 
\begin{tabularx}{\linewidth}{>{\hsize=0.38\hsize\linewidth=\hsize}X>{\hsize=0.37\hsize\linewidth=\hsize}X>{\hsize=0.30\hsize\linewidth=\hsize}X}
\toprule
\textbf{Task} & \textbf{Adapted from:} & \textbf{Number of Items}\\ \midrule
Sentence Completion & \textbf{BVL} \citep{BVL} & 14 items \\ 
Acceptability Judgment & \textbf{BVL} \citep{BVL} & 18 items \\ 
Idiom Comprehension & \textbf{BVL} \citep{BVL} & 10 items \\ 
Sentence Comprehension & \textbf{BVL} \citep{BVL} & BVL: 40 items\\
Sentence Comprehension & \textbf{TROG-2} \citep{TROG-2}  & TROG-2: 80 items \\Sentence Comprehension & \textbf{TCGB-2} \citep{TCGB-2} & TCGB-2: 74 items \\
Lexical Comprehension & \textbf{BVL} \citep{BVL} & BVL: 18 items\\
Lexical Comprehension & \textbf{PPVT-R} \citep{Peabody_it}\footnotemark{} & PPVT-R: 165 items \\
\bottomrule
\end{tabularx}
}
\caption{BaBIEs: General Structure}
\label{tab:babies_structure}
\end{table}

\footnotetext{The original version of the Peabody test contains 175 items. Ten items were excluded during the adaptation process, because either the words were too uncommon or it was impossible to convert them into linguistic expressions.}

Each task in BaBIEs targets different aspects of linguistic competence, thus providing a global linguistic profile of the models. Furthermore, BaBIEs is particularly suited for a comprehensive evaluation of the BabyLMs' syntactic competence, as each Sentence Comprehension task addresses partially distinct syntactic structures (e.g., BVL: Reflexive Active clauses, Agreement, Clitic, etc.; TROG-2: Reversible ‘in’ and ‘on’, Pronoun Binding, Zero anaphor, etc.; TCGB-2: Locative clauses, Dative clauses, etc.). 

The Sentence Completion Task is the only one addressing linguistic production. Namely, it assesses the ability to produce verb inflected forms, with a focus on number and tense morphology. Each item is an incomplete sentence (e.g., \textit{Il papà parte spesso per lavoro. Anche ieri il papà <mask>} `Dad often travels for work. Even yesterday, Dad <mask>'). The model must provide the (target) answer (e.g., \textit{è partito, partiva} `left, was leaving') through a fill-in-the-blank task. In this case, \textit{beam search} was selected as generation strategy, with 3 beams. Each response was scored as correct (cf. Section \ref{subsect_metrics} below).

The Acceptability Judgment Task contains 18 minimal pairs of sentences. In order to obtain minimal pairs, we created a grammatical/ungrammatical version of the original sentence, depending on its (un)grammaticality (e.g., original sentence: \textit{La mela è rossa} `The apple is red.SING'; ungrammatical version: \textit{*La mela è rosse} `*The apple is red.PLUR'). 

The items in the tasks specifically targeting comprehension (i.e., Idioms, Sentence, and Lexical Comprehension tasks) follow a similar structure, consisting of one linguistic stimulus plus a set of possible answers. An example item for each task is provided in Table \ref{tab:examples_comprehension_tasks}. The linguistic stimuli and each possible answer are concatenated through the conjunction \textit{cioè} `that is', so as to create four sentences, from which the models must select the target one. The models' selection is based on the perplexity score they assign to each sentence. 

\begin{table}[!h]
\centering
\resizebox{\textwidth}{!}{ 
\begin{tabularx}{\linewidth}{>{\hsize=0.4\hsize\linewidth=\hsize}X>{\hsize=0.6\hsize\linewidth=\hsize}X}
\toprule
Linguistic Stimulus & Set of possible answers and \textbf{Target answer}\\ \midrule \\
Quel ragazzo si dà delle arie \newline `That boy puts on airs' \newline (Idiom Comprehension) & 1. Cioè quel ragazzo fa finta di niente \newline `That is, that boy acts as nothing is wrong' \newline 2. Cioè quel ragazzo respira \newline `That is, that boy is breathing' \newline \textbf{3. Cioè quel ragazzo cerca di apparire importante} \newline \textbf{`That is, that boys is trying to seem important'} \\ \\
Il cane è tirato dall'uomo \newline `The dog is pulled by the man' \newline (Sentence Comprehension) & 1. Cioè il cane tira l'uomo \newline `That is, the dog pulls the man' \newline 2. Cioè l'uomo tiene il cane \newline `That is, the man holds the dog' \newline 3. Cioè l'uomo chiama il cane \newline `That is, the man calls the dog' \newline \textbf{4. Cioè l'uomo tira il cane} \newline \textbf{`That is, the man pulls the dog'} \\ \\
Un balcone \newline `A balcony' \newline (Lexical Comprehension) & \textbf{1. Cioè un terrazzino} \newline \textbf{`That is, a small roof'} \newline 2. Cioè una fontana \newline `That is, a fountain' \newline 3. Cioè un portico \newline `That is, a porch' \newline 4. Cioè un portone \newline `That is, a front door' \\
\bottomrule
\end{tabularx}}
\caption{BaBIEs: Example items of comprehension tasks}
\label{tab:examples_comprehension_tasks}
\end{table}

It is worth mentioning that although the items reported in Table \ref{tab:examples_comprehension_tasks} have a similar structure in BaBIEs, their original versions are different. The sets of possible answers were linguistic expressions in the case of the Idiom Comprehension Task, whilst they were pictures in the case of both the Sentence and the Lexical Comprehension Tasks. Hence, for these two tasks, the sets of possible answers are the result of a picture-to-language conversion process. Regarding the Sentence Comprehension Tasks, each target answer is a sentence that differs from its stimulus syntactically, but not lexically. Conversely, in the Lexical Comprehension tasks, the target answers can be either sentences, phrases or nouns. In the latter case, the target answer can be a synonym, meronym or hyponym of the stimulus (for more details, see \cite{capone2024babies}).

\subsection{Metrics} \label{subsect_metrics}

The performance of the models is evaluated using two closely related metrics: i.) accuracy and ii.) age-equivalent scores. Accuracy is a direct measure of the model's performance, while age-equivalent scores also take into account the size of the training dataset for the evaluation. Combining these metrics allows for a comprehensive assessment of the models' linguistic abilities and facilitates meaningful comparisons with children and among models trained with different resources.

Accuracy, which measures the proportion of correct predictions or target answers relative to the total number of items is a widely used metric for evaluating LMs. It also serves as the basis for determining age-equivalent scores in child evaluations. These scores are assigned based on the combination of the accuracy reached by the child in a given task and the child age. The procedure for assigning age-equivalent scores in standardized tests operates as follows: first, the child's raw accuracy score for a given task is calculated as the ratio of target responses out of the total number of items. This score is then converted into an age-equivalent score based on the child's age.
To illustrate, consider two children, aged 3;6 and 4;6 years old, respectively, who both achieve the same accuracy-score of 65 in a Lexical Comprehension Task. While their accuracy scores are identical, the age-equivalent score will be higher for the younger child and lower for the older one, aligning with their respective developmental stages developmental stages. 
Age-equivalent scores are determined using the standardization sample as a reference point and are interpreted relative to it. Since the score distribution of the standardization sample is normal, it is possible to assess whether the child's age-equivalent score falls within the typical range: $\pm$ 2.5 SD (Standard Deviation) from the average score for their reference age range. Additionally, it is possible to infer the child's linguistic age by identifying the age at which their age-equivalent score aligns with the average score or corresponds to the 50th percentile. 

The raw scores of the models are calculated according to the specific procedures outlined in each test, which vary slightly. For the BVL and the Peabody, the raw score corresponds simply to the number of target responses.\footnote{For the Peabody test, age-equivalent scores are calculated based on 175 items. To account for the excluded items, a 10-item raw-score range is used to determine the models' age-equivalent scores.} For the TCGB-2, however, the raw score is based on the error score, making it the inverse of accuracy. In this test, each error is assigned a score of $0.5$, if the participant selects an incorrect answer once, and a score of $1$, if the incorrect answer is selected twice. Notably, in child evaluations, the experimenter is allowed to repeat the question if the target answer is not provided on the first attempt. For the evaluation of the models, as none of the questions were repeated, each error was consistently assigned a score of $0.5$. Finally, for the TROG-2, items are grouped into 20 four-item blocks, with a block being considered ``passed'' if at least three out of four items are answered correctly. The raw score is then determined by the total number of passed blocks. 

Before addressing the age-equivalent scores, it is necessary to clarify an important point about the Sentence Completion Task. As noted in the previous section, this task is designed to assess knowledge of verb inflection. However, the target responses are required not only to be morphologically and syntactically correct, but also to be semantically appropriate.
As will be illustrated in the next section, models often produce responses that are morphologically and syntactically correct, while failing to meet semantic appropriateness. Therefore, two scoring procedures are adopted for this task: the first considers only responses that are both syntactically correct and semantically appropriate as correct (\emph{Strict Scoring}), while the second also includes syntactically correct but semantically inappropriate responses as correct (\emph{Loose Scoring}). For example, consider the stimulus \textit{La mamma cucina. Le mamme <mask>} `Mommy is cooking. Mommies <mask>'. Under the Strict Scoring, the only correct response is \textit{cucinano} `are cooking'. Under the Loose Scoring, a response like \textit{e i papà faranno la} `and daddies will make the' is scored as correct, since the verb is correctly inflected for the third plural person, as requested by the task, even if the sentence is not actually completed. Under both scoring procedure, answers that do not contain a verb are scored as incorrect.

To determine the age-equivalent scores for the models, each must be assigned an age. We defined the \emph{model age} in terms of the number of word tokens used for its training. Consequently, Minerva and SmolVLM are treated as ``adults'', meaning they are evaluated using the highest age range considered in each test as a reference. Conversely, the four BAMBI models are treated as being in the 5;0–5;5 age range. 

Finally, in this study, scores within the range of $\pm$ 1 standard deviation (SD) from the average score— or the corresponding percentiles — are considered ``typical'', differently from acquisitional studies, which often classify scores as atypical when they fall outside the range of $\pm$ 2.5 SD from the average.

\subsection{Results} \label{subsect_results}

\begin{figure} [!h]
\begin{centering}
\includegraphics[width=1\linewidth, trim = {0, 2.5cm, 0, 3cm}, clip] {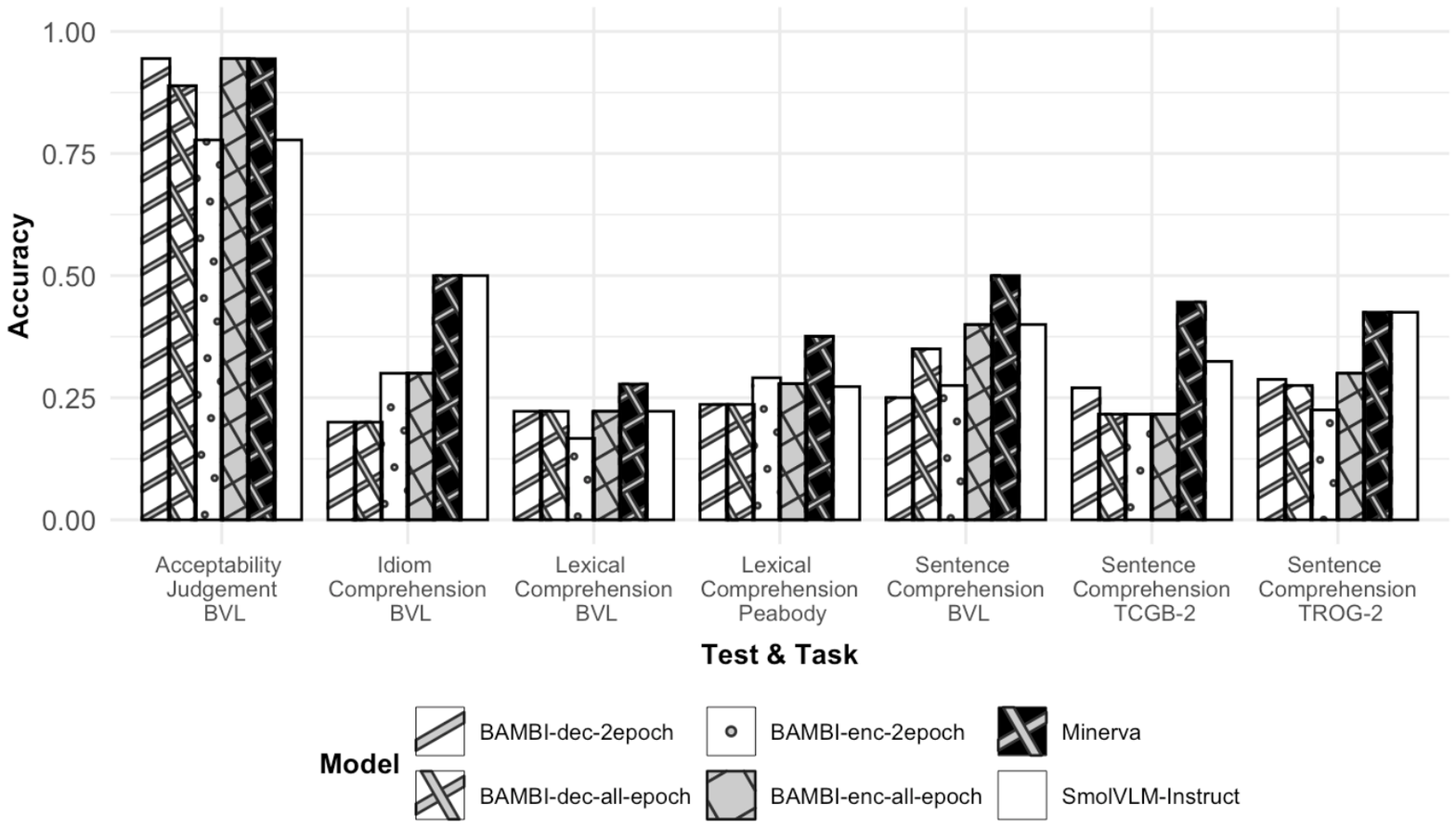}
\par\end{centering}
\caption{\label{figure_1}\textsc{Accuracy reached by the models. Comprehension and Acceptability Judgment tasks}.}
\end{figure}

The accuracy achieved by the models in the Comprehension and Acceptability Judgment tasks is illustrated in Figure \ref{figure_1}, while the age-equivalent scores are reported in Table\ref{tab:agequiv_all_models}. The results of the Sentence Completion task are treated separately, due to the twofold scoring procedure introduced in Section \ref{subsect_metrics}.

\begin{table}[!h]
\centering
\resizebox{\textwidth}{!}{ 
\begin{tabular}{@{}lllllll@{}}
\toprule
\textbf{} & \specialcell[t]{\textbf{BAMBI} \\ \textbf{dec-2epoch}} & \specialcell[t]{\textbf{BAMBI} \\ \textbf{dec-all-epoch}} & \specialcell[t]{\textbf{BAMBI} \\ \textbf{enc-2epoch}} & \specialcell[t]{\textbf{BAMBI} \\ \textbf{enc-all-epoch}} & \textbf{Minerva} & \textbf{SmolVLM} \\ \midrule
\specialcell[t]{\textbf{Acceptability J.} \\ \textbf{(BVL)}} & \specialcell[t]{+2SD \\ (5;0-5;5)} & \specialcell[t]{+2SD \\ (5;0-5;5)} & \specialcell[t]{+2SD \\ (5;0-5;5)} & \specialcell[t]{+2SD \\ (5;0-5;5)} & \specialcell[t]{-2SD \\ (11;6-11;11)} & \specialcell[t]{-2SD \\ (11;6-11;11)} \\
\specialcell[t]{\textbf{Idiom C.} \\ \textbf{(BVL)}} & \specialcell[t]{-1SD<x<0 \\ (5;0-5;5)} & \specialcell[t]{-1SD<x<0 \\ (5;0-5;5)} & \specialcell[t]{0<x<+1SD \\ (5;0-5;5)} & \specialcell[t]{0<x<+1SD \\ (5;0-5;5)} & \specialcell[t]{-2SD \\ (11;6-11;11)} & \specialcell[t]{-2SD \\ (11;6-11;11)} \\
\specialcell[t]{\textbf{Lexical C.} \\ \textbf{(BVL)}} & \specialcell[t]{-2SD \\ (5;0-5;5)} & \specialcell[t]{-2SD \\ (5;0-5;5)} & \specialcell[t]{-2SD \\ (5;0-5;5)} & \specialcell[t]{-2SD \\ (5;0-5;5)} & \specialcell[t]{-2SD \\ (11;6-11;11)} & \specialcell[t]{-2SD \\ (11;6-11;11)} \\
\specialcell[t]{\textbf{Lexical C.} \\ \textbf{(PPVT)}} & \specialcell[t]{Below average \\ (4;9-5;6)} & \specialcell[t]{Below average \\ (4;9-5;6)} & \specialcell[t]{Below average \\ (4;9-5;6)} & \specialcell[t]{Below average \\ (4;9-5;6)} & \specialcell[t]{Below average \\ (10;7-11;6)} & \specialcell[t]{Below average \\ (10;7-11;6)}\\
\specialcell[t]{\textbf{Sentence C.} \\ \textbf{(BVL)}} & \specialcell[t]{-2SD \\ (5;0-5;5)} & \specialcell[t]{-2SD \\ (5;0-5;5)} & \specialcell[t]{-2SD \\ (5;0-5;5)} & \specialcell[t]{-2SD \\ (5;0-5;5)} & \specialcell[t]{-2SD \\ (11;6-11;11)} & \specialcell[t]{-2SD \\ (11;6-11;11)} \\
\specialcell[t]{\textbf{Sentence C.} \\ \textbf{(TCGB-2)}} & \specialcell[t]{Equivalent Ling. age: \\ 3;6-4;5} & \specialcell[t]{Equivalent Ling. age: \\ 3;6-4;5} & \specialcell[t]{Equivalent Ling. age: \\ 3;6-4;5} & \specialcell[t]{Equivalent Ling. age: \\ 3;6-4;5} & \specialcell[t]{Equivalent Ling. age: \\ 3;6-4;5} & \specialcell[t]{Equivalent Ling. age: \\ 3;6-4;5} \\
\specialcell[t]{\textbf{Sentence C.} \\ \textbf{(TROG-2)}} & \specialcell[t]{Equivalent Ling. age: \\ < 4;2} & \specialcell[t]{Equivalent Ling. age: \\ < 4;2} & \specialcell[t]{Equivalent Ling. age: \\ < 4;2} & \specialcell[t]{Equivalent Ling. age: \\ < 4;2} & \specialcell[t]{Equivalent Ling. age: \\ 5;0} & \specialcell[t]{Equivalent Ling. age: \\ 4;2} \\
\bottomrule
\end{tabular}
}
\caption{Age-equivalent scores and equivalent linguistic ages (both expressed in years; month). Acceptability Judgment and Comprehension tasks, all models}
\label{tab:agequiv_all_models}
\end{table}

All models achieve their highest accuracy in the Acceptability Judgment Task (BVL). As reported in Figure \ref{figure_1}, Minerva, BAMBI-dec-2epoch, and BAMBI-enc-all-epoch reach to values. The accuracy obtained by BAMBI-dec-all-epoch and BAMBI-enc-2epoch is slightly lower. The lowest score is obtained by SmolVLM. 
Let us now turn to the age-equivalent scores. Since all BAMBI models are assigned a "model age" of 5 years, their performance is evaluated within the 5;0–5;5 age range, and their scores exceed +2 SD for this group. Minerva and SmolVLM, on the other hand, are evaluated using the highest age range considered by the BVL, i.e., 11;6–11;11 years, for which Minerva's score falls between -1 SD, whilst that of SmolVLM below -2 SD. Concerning this task, the equivalent linguistic age of SmolVLM is 4;0-4;6 (since its score falls within $\pm$ 1 SD for this age range).

The accuracy reached by the models drops markedly in all the other tasks. In the Idiom Comprehension task, the two decoder versions of BAMBI achieve the lowest accuracy. The accuracy achieved by the encoder versions is slightly higher, while the highest accuracy is obtained by both Minerva and SmolVLM. However, if we also consider the age-equivalent scores, a different picture emerges. The score achieved by the decoder versions of BAMBI falls between -1 DS and 0 for the 5;0-5;5 age range. The score of the two encoder versions falls between 0 and +1 SD for the same age range. In contrast, Minerva's and SmolVLM's scores fall below -2 SD for their reference age range and would fall within $\pm$1 SD for the 7;6-7;11 age range.

The Lexical Comprehension tasks appear to be the most challenging comprehension task for the models. In the task derived from BVL, none of the models achieve an accuracy higher than 0.30. Minerva attains the highest accuracy, followed by the two decoder versions of BAMBI, BAMBI-enc-all-epoch, and SmolVLM, which all achieve the same accuracy. BAMBI-enc-2epoch records the lowest accuracy. All the models' age-equivalent scores fall well below -2 SD from the average for the reference age range, as well as -2 SD from the average for the lowest age-range considered by the test (4;0-4;5 years). 
The same goes for the task from Peabody, where Minerva performs better than the other models, but still achieves an age-equivalent score below average for its reference age-range (i.e., 10;7-11;6). In this task, the second highest accuracy is obtained by BAMBI-enc-2epoch, followed by SmolVLM and BAMBI-enc-all-epoch. The accuracy achieved by the two decoder versions of BAMBI is slghtly lower. All their age-equivalent scores fall below average for their respective reference age ranges (4;9–5;6 years for the BAMBI models and 10;7–11;6 years for SmolVLM). 

Let us now turn to the Sentence Comprehension task. All models perform better than in the Lexical Comprehension one, and yet they all struggle to achieve an accuracy beyond 0.50. Minerva consistently achieves the highest accuracy, followed by SmolVLM. BAMBI-anc-all-epoch obtains the same accuracy as SmolVLM in the task from BVL. It performs better than all the other versions of BAMBI also in the task from TROG-2, immediately followed by BAMBI-dec-2-epoch. For the task from BVL, all the age-equivalent scores obtained by the models fall below -2 SD for their respective reference age ranges (5;0–5;5 years for BAMBI, 11;6–11;11 years for Minerva and SmolVLM). However, while the scores of Minerva and SmolVLM fall between -1 SD and 0 for the 4;0–4;5 age range (the minimum age considered by the test), the scores of all BAMBI models remain below -2 SD even for this age range. As for the TROG-2, according to their age-equivalent scores, Minerva has a linguistic age of 5 years, SmolVLM is 4;2 years old, while the equivalent linguistic age of BAMBI models is below 4;2 years. Finally, all models have an equivalent linguistic age between 3;6 and 4;5 years, regarding the task from TCGB-2.

\begin{figure} [!h]
\begin{centering}
\includegraphics[width=1\linewidth, trim = {0, 2.5cm, 0, 3cm}, clip] {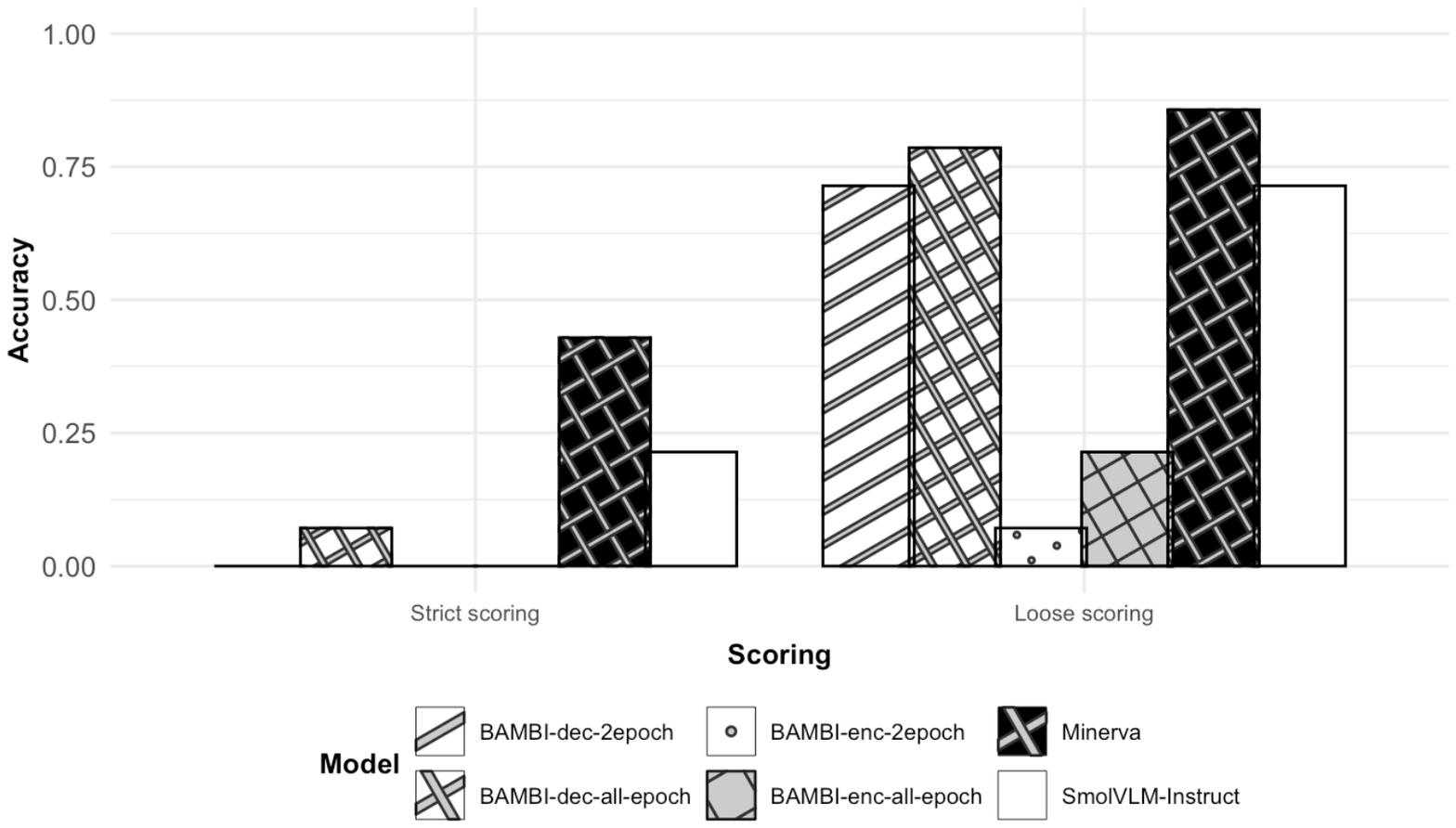}
\par\end{centering}
\caption{\label{figure_2}\textsc{Accuracy reached by the models. Sentence Completion Task, strict and loose scoring}.}
\end{figure}

Finally, we present the accuracy achieved by the models in the sole task addressing linguistic production -- the Sentence Completion Task. As outlined in Section \ref{subsect_metrics}, two scoring procedures are applied to the responses generated by the models: the Strict and the Loose Scooring. The results are illustrated in Figure \ref{figure_2}. Table \ref{tab:accuracy_all_models_sentence_completion} summarizes the accuracy values obtained by the models.

\begin{table}[!h]
\centering
\resizebox{\textwidth}{!}{ 
\begin{tabularx}{\linewidth}{>{\hsize=0.40\hsize\linewidth=\hsize}X>{\hsize=0.30\hsize\linewidth=\hsize}X>{\hsize=0.30\hsize\linewidth=\hsize}X}
\toprule
\textbf{Models/Scoring Method} & \textbf{Strict Scoring} & \textbf{Loose Scoring} \\ \midrule
\textbf{BAMBI-dec-2epoch} & 0.00  & 0.71 \\
\textbf{BAMBI-dec-all-epoch} & 0.07  & 0.78 \\
\textbf{BAMBI-enc-2epoch} & 0.00  & 0.07 \\
\textbf{BAMBI-enc-all-epoch} & 0.00  & 0.21 \\
\textbf{Minerva} & 0.42  & 0.86 \\
\textbf{SmolVLM-Instruct} & 0.21  & 0.71 \\
\bottomrule
\end{tabularx}}
\caption{Accuracy values. Sentence Completion Task, Strict and Loose Scoring, all models}
\label{tab:accuracy_all_models_sentence_completion}
\end{table}

The Sentence Completion task is challenging overall, especially for the BAMBI models, while Minerva demonstrates the best performance. Under the Strict Scoring (where only responses that are both syntactically correct and semantically appropriate are considered correct), all models exhibit low accuracies, with age-equivalent scores falling below -2 SD for the minimum age considered by the test (4;0 years).

In contrast, considering the Loose Scoring (where a response is scored as correct if it is syntactically correct), all models achieve higher accuracies and improved age-equivalent scores. The age-equivalent scores of the two decoder versions of BAMBI exceed +1 SD (dec-2epoch) and +1.5 SD (dec-all-epoch) for their reference age range (5;0–5;5 years). SmolVLM shifts to an age-equivalent score below -2 SD for its reference age range (11;11 years) but falls within the range of +1 SD to -1 SD for the age range of 5;0–9;5 years. Minerva's age-equivalent score ranges between -1.5 SD and -1 SD for its reference age range (11;11 years) and between -1 SD and 0 for the age range of 8;6–9;5 years.

Finally, the accuracy achieved by the two encoder versions of BAMBI does not significantly improve using the Loose Scoring. Their age-equivalent scores remain below -2 SD for the minimum age considered by the test.

\subsection{Discussion} \label{subsect_discussion}

The evaluation of the BAMBI models using BaBIEs reveals that the BabyLMs, despite being trained for only a few epochs on a limited dataset, display an equivalent linguistic age ranging approximately from 3;6 to 5;5 years, depending on the task. Notably, BabyLMs trained with the unrestricted strategy (i.e., without limiting training to two epochs) do not demonstrate significant improvements over those trained with the restricted approach.
When comparing their performance to that of Minerva and SmolVLM, two key aspects emerge. First, the BAMBI models outperform larger models in specific tasks. For instance, in the Acceptability Judgment Task two BAMBI models reach higher accuracies than SmolVLM. 
Even in tasks in which they actually achieve overall lower accuracies, such as the Sentence Comprehension tasks, this trend does not always hold when specific syntactic structures are considered (see table \ref{tab:accuracy_neg_all_models}). A notable example is negation, which is represented by seven different structures in BaBIEs (\ref{ex_1}):

\begin{exe}[(9)]
\fontsize{10}{13}\selectfont
  \ex \label{ex_1} \begin{xlist}
   \ex[]{\textbf{Double Negation}: Né il bambino né la bambina mangiano `Neither the boy nor the girl eats' 
    \glt Target answer: Cioè il bambino non mangia e la bambina non mangia `That is, the boy does not eat and the girl does not eat}\label{ex_1a} 
    \ex[]{\textbf{Negative Active clause}: Il cane non corre `The dog does not run'
    \glt Target answer: Cioè il cane è seduto vicino al gatto `That is, the dog sits next to the cat'}\label{ex_1b}
    \ex[]{\textbf{Negative Passive clause}: La macchina non è lavata dal bambino `The car is not washed by the child'
    \glt Target answer: Cioè il papà lava la macchina e il bambino guarda il papà `That is, the dad washes the car, and the child watches the dad'}\label{ex_1c} 
    \ex[]{\textbf{Reversible Negative Passive clause}: Il cane non è seguito dal gatto `The dog is not followed by the cat'
    \glt Target answer: Cioè il cane segue il gatto `That is, the dog follows the cat'}\label{ex_1d} 
    \ex[]{\textbf{Negation}: La stella non è rossa `The star is not red'
    \glt Target answer: Cioè la stella è di colore bianco `That is, the star is white'}\label{ex_1e} 
     \ex[]{\textbf{Not only X but Y}: La matita non è soltanto lunga ma anche rossa `The pencil is not only long but also red'
    \glt Target answer: Cioè la matita è rossa ed è lunga `That is, the pencil is red and it is long'}\label{ex_1f} 
    \ex[]{\textbf{X but not Y}: L'uomo, ma non il cavallo, sta saltando `The man, but not the horse, jumps'
    \glt Target answer: Cioè l'uomo salta e il cavallo è fermo `That is, the man jumps, and the horse stands still'}\label{ex_1g} 
  \end{xlist}
\end{exe}

\noindent{}The accuracy achieved by the models for the structures in (\ref{ex_1a}-\ref{ex_1g}) is shown in Table \ref{tab:accuracy_neg_all_models}. 

\begin{table}[!h]
\centering
\resizebox{\textwidth}{!}{ 
\begin{tabular}{@{}lcccccc@{}}
\toprule
\textbf{} & \specialcell[t]{\textbf{BAMBI} \\ \textbf{dec-2epoch}} & \specialcell[t]{\textbf{BAMBI} \\ \textbf{dec-all-epoch}} & \specialcell[t]{\textbf{BAMBI} \\ \textbf{enc-2epoch}} & \specialcell[t]{\textbf{BAMBI} \\ \textbf{enc-all-epoch}} & \textbf{Minerva} & \textbf{SmolVLM} \\ \midrule
\textbf{Double Negation} & 0.67  & 0.50 & 0.67 & 0.67 & 0.33 & 0.33 \\
\textbf{Negative Active} & 0.10  & 0.30 & 0.10 & 0.40 & 0.40 & 0.25 \\
\textbf{Negative Passive} & 0.50  & 0.37 & 0.25 & 0.12 & 0.25 & 0.25 \\
\textbf{Revers. Negative Passive} & 0.00  & 0.00 & 0.00 & 0.00 & 0.00 & 0.00 \\
\textbf{Negation} & 0.50  & 0.50 & 0.00 & 0.50 & 0.25 & 0.50 \\
\textbf{Not only X but Y} & 0.00  & 0.50 & 0.50 & 0.50 & 0.25 & 0.50 \\
\textbf{X but not Y} & 0.25  & 0.25 & 0.00 & 0.50 & 0.25 & 0.25 \\
\bottomrule
\end{tabular}
}
\caption{Accuracy values for the syntactic structures involving negation, all models}
\label{tab:accuracy_neg_all_models}
\end{table}

For six out of seven negative structures, at least one version of BAMBI models achieves the highest accuracy. For Double Negation, Negative Passive clauses, and X but not Y, the accuracy obtained by BAMBI models is higher than those obtained by Minerva and SmolVLM. 
While accuracy offers a valuable metric for comparing models' performances, it lacks informative features about their training process. However, the BaBIEs benchmark provides an evaluation metric that incorporates training-related features, that is age-equivalent scores. In certain tasks, using accuracy alone may be misleading and age-equivalent scores must complement it. For example, in the Idiom Comprehension Task, the age-equivalent scores of BAMBI models surpass those of both Minerva and SmolVLM, despite having lower accuracies (see Figure\ref{figure_1}).

The second key aspect is that the behavior BAMBI models exhibit is similar to that of Large LMs, such as Minerva and SmolVLM.
Namely, the BAMBI models and the two larger LMs under investigation perform better on syntactic tasks but face greater challenges with semantic tasks. This is evidenced by the higher accuracies achieved in the Sentence Comprehension tasks compared to the Lexical Comprehension tasks. In the latter, the primary challenge lies in the semantic relationship between the stimulus and the target answer (e.g., synonymy, hyponymy, paraphrasis, or meronymy), whereas, in the Sentence Comprehension Task, each stimulus differs from its target only syntactically, without any semantic or lexical variation. 
This discrepancy between semantics and syntax is also confirmed by the Sentence Completion Task, where the models (excluding the two encoder versions of the BAMBI models) tend to generate responses which are syntactically and morphologically correct, but not semantically appropriate, as highlighted by the two scoring procedures, that yielded very different accuracy values. 
The observation that the BAMBI models exhibit stronger syntactic competence alongside limited semantic understanding aligns not only with the behavior displayed by the larger LLMs in this study but also with prior findings \citep{zhang2020you}.
Furthermore, the nature of the task (such as generation versus probability based evaluation) plays an additional role in shaping model performance, as evidenced by the mismatch in the accuracy all models achieve in the Acceptability Judgment Task, compared to those achieved in all of the other tasks. This outcome is expected, since the task uses minimal sentence pairs, which have already proven to be efficient in evaluating the syntactic competence of LMs, and it also highlights how the behavior of the BAMBI models aligns to those of larger LMs.

The results, expressed in terms of age-equivalent scores, show that increasing the size of the model architecture, dataset, and computational resources does not produce models with equivalent age scores proportional to the input received. Specifically, the BAMBI decoder models demonstrate greater consistency in their age-equivalent scores compared to the pre-trained models examined. Moreover, the results appear to confirm the findings of the BabyLM Challenges \citep{warstadt2023findings, hu2024findings}, which suggest that multimodal LMs does not outperform classical language modeling as a training strategy. This suggests that the multimodal information embedded in current LMs is still not enough to make a difference in modeling the role of sensorimotor experience in language acquisition.

\section{Conclusions} \label{sect_conclusions}

In this study, we presented BAMBI, a series of Italian BabyLMs trained on a dataset that closely mirrors the linguistic input received by a 5 y.o. child, from both a quantitative and a qualitative perspective. 
The models are evaluated against two Large LMs, Minerva and SmolVLM (multimodal). For the evaluation, we used BaBIEs, a benchmark specifically designed to evaluate BabyLMs in Italian. 
Overall, the BAMBI models demonstrated quite robust syntactic competence, regardless of the amount of linguistic input.

Even if their performances lag behind that of Minerva and SmolVLM in some of the tasks,  the observed differences in accuracy are not directly proportional to the amount of training data (either linguistic or multimodal) or computational resources. Such conclusion is further strengthened by the age-equivalent score metric. This suggests room for performance improvements through strategies other than simply scaling dataset size. These findings represent the initial phase of a broader training curriculum designed to incorporate additional training subsets, which builds upon a core hypothesis which stipulates that ``starting small'' with simpler data establishes stronger grounding for subsequent training \citep{elman1993learning}.

\section*{Acknowledgments}
We acknowledge financial support under the PRIN 2022 Project Title "Computational and linguistic benchmarks for the study of verb argument structure" – CUP I53D23004050006 - Grant Assignment Decree No. 1016 adopted on 07/07/2023 by the Italian Ministry of University and Research (MUR). This research was also partly funded by PNRR—M4C2—Investimento 1.3, Partenariato Esteso PE00000013—“FAIR—Future Artificial Intelligence Research”—Spoke 1 “Human-centered AI,” funded by the European Commission under the NextGeneration EU programme.

\authorinfo{Alice Suozzi}{Ca' Foscari University of Venice}{Ca' Bembo, Fondamenta Tofetti, Dorsoduro 1075 - Venice}{Italy}{alice.suozzi@unive.it}
\authorinfo{Luca Capone}{University of Pisa}{via Santa Maria 36 - Pisa}{Italy}{luca.capone@fileli.unipi.it}
\authorinfo{Gianluca E. Lebani}{Ca' Foscari University of Venice}{Ca' Bembo, Fondamenta Tofetti, Dorsoduro 1075 - Venice}{Italy}{gianluca.lebani@unive.it}
\authorinfo{Alessandro Lenci}{University of Pisa}{via Santa Maria 36 - Pisa}{Italy}{alessandro.lenci@unipi.it}





\end{document}